%% file: main.tex
\documentclass[letterpaper]{article} 
\usepackage{aaai24}  
\usepackage{times}  
\usepackage{helvet}  
\usepackage{courier}  
\usepackage[hyphens]{url}  
\usepackage{graphicx} 
\usepackage{natbib}  
\usepackage{caption} 
\usepackage[utf8]{inputenc} 
\usepackage[T1]{fontenc}    
\usepackage{booktabs}       
\usepackage{amsfonts}       
\usepackage{nicefrac}       
\usepackage{microtype}      
\usepackage{xcolor}         
\usepackage{float}
\usepackage{longtable}
\usepackage{fancyvrb} 
\usepackage{graphicx}
\usepackage{caption}
\usepackage{subcaption}
\usepackage{microtype}
\usepackage{algorithm}
\usepackage{algpseudocode}
\usepackage[export]{adjustbox}

\title{How much can change in a year? Revisiting Evaluation in Multi-Agent Reinforcement Learning}
\author {
    Siddarth Singh\textsuperscript{\rm 1},
    Omayma Mahjoub\textsuperscript{\rm 1},
    Ruan de Kock\textsuperscript{\rm 1},
    Wiem Khlifi\textsuperscript{\rm 1, \rm 2},
    Abidine Vall\textsuperscript{\rm 3},
    Kale-ab Tessera\textsuperscript{\rm 4},
    Arnu Pretorius\textsuperscript{\rm 1}
}
\affiliations {
    \textsuperscript{\rm 1}InstaDeep Ltd\\
    \textsuperscript{\rm 2}National School of Computer Science, Tunisia\\
    \textsuperscript{\rm 3}National School of Engineering of Tunis\\
    \textsuperscript{\rm 4}University of Edinburgh\\
}

\begin{document}

\maketitle
\input{sections/0-abstract}
\input{sections/1-introduction}
\input{sections/2-benchmarks}
\input{sections/3-Trends}
\input{sections/4-solutions}
\input{sections/5-Conclusion}

\bibliography{Indaba}

\end{document}

%% file: sections/0-abstract.tex
\begin{abstract}
  Establishing sound experimental standards and rigour is important in any growing field of research. Deep Multi-Agent Reinforcement Learning (MARL) is one such nascent field. Although exciting progress has been made, MARL has recently come under scrutiny for replicability issues and a lack of standardised evaluation methodology, specifically in the cooperative setting. Although protocols have been proposed to help alleviate the issue, it remains important to actively monitor the health of the field. In this work, we extend the database of evaluation methodology previously published by \citep{gorsane2022standardised} containing meta-data on MARL publications from top-rated conferences and compare the findings extracted from this updated database to the trends identified in their work. Our analysis shows that many of the worrying trends in performance reporting remain. This includes the omission of uncertainty quantification, not reporting all relevant evaluation details and a narrowing of algorithmic development classes. Promisingly, we do observe a trend towards more difficult scenarios in SMAC-v1, which  if continued into SMAC-v2 \citep{ellis2022smacv2} will encourage novel algorithmic development. Our data indicate that replicability needs to be approached more proactively by the MARL community to ensure trust in the field as we move towards exciting new frontiers. 
\end{abstract}

%% file: sections/1-introduction.tex
\section{Introduction}

Multi-Agent Reinforcement Learning (MARL) is a rapidly growing field that has gained more attention in recent years, with large amounts of new algorithms in development which have shown the ability to solve complex problem solving tasks like Starcraft \citep{SMAC}, Hanabi \citep{foerster2019bayesian,hu2021simplified} and recently Diplomacy by integrating new developments from other fields like LLMs \citep{Bakhtin2022HumanlevelPI}. Given these successes there is a growing interest in utilising MARL to solve real-world problems like resource allocation, management and sharing, network routing and traffic signal controls \citep{IntelligentTrafficControl, brittain2019autonomous, DynamicPowerAllocation, AirTrafficManagement, IoTNetworksWithEdgeComputing, TrajectoryDesignandPowerAllocation, pretorius2020game, InternetofControllableThings} as these problems naturally lend to being formulated as cooperative MARL systems where a team of agents coordinates to optimise a shared global reward. However, similar to other fields within machine learning, ensuring a high level of scientific rigour and sound experimental methodology has become difficult as algorithms become more complex and computational requirements more extreme. As we move away from synthetic problems like video games towards tasks with real-world repercussions, it is important to ensure that trust in the field as a whole is not eroded over time by limited explainability of our results. Fortunately, these issues have been identified as they appeared and multiple works have suggested approaches to improve this and identify areas of concern before they become too widespread \citep{colas2018random, henderson2019deep, colas2019hitchhikers, engstrom2020implementation, jordan2020evaluating, agarwal2022deep}.
An important first step in encouraging trust in the field of MARL is clarity and an accessible means for people to be able to determine the health of the field without requiring individual data gathering. To this end  \cite{gorsane2022standardised} compiled a dataset of evaluation methodologies used in deep cooperative MARL, from the first paper published in January 2016 up until April 2022. This database is open source and with only a small amount of database manipulation, it is possible on the reader's side to evaluate the health of MARL and determine popular algorithms, settings and interesting new avenues for development. Importantly, it also allows readers to determine which publications have made a strong effort to ensure high rigour and replicability. The authors discovered some worrying trends in research w.r.t. the consistency of reported results. As the number of publications grows in the coming years, it is important to update this dataset with new publications and perform a continuous assessment of the field. Although the original analysis of \citep{gorsane2022standardised} highlighted many issues, there was evidence of a more thorough evaluation beginning to gain traction in more recent years, with an increase in the number of ablation studies and an upward trend in the use of uncertainty quantification. 

In this paper, we treat the original dataset compiled by \citep{gorsane2022standardised} as the historical trends of MARL evaluation and compare this to recent developments since the original publication. We compile a new dataset from the period of April 2022 to December 2022 which covers 29 papers. These papers are sourced under the same standards as the historical dataset and only include works published at high tier conferences including NeurIPS, ICLR and ICML with a focus on deep cooperative MARL. Using this dataset of recent trends, we show in which areas evaluation methodology has improved or stagnated, highlight changes in algorithmic and environment popularity and provide recommendations\footnote{The updated dataset is available at the following \url{https://drive.google.com/file/d/1psSUrS8ywPAJY1Qm4ySfTYcYYJs0TlLF}}.

%% file: sections/2-benchmarks.tex
\section{Background}

\textbf{Algorithm Training. } In cooperative MARL, the majority of algorithms can broadly be split into 2 paradigms; Centralised Training Decentralised Execution (CTDE) and Decentralised Training Decentralised Execution (DTDE). DTDE methods like Independent Q-Learning \citep{tan1993multi} train agents using a policy based purely on their own egocentric experiences in the training environment. These methods are often called Independent Learners (IL) \citep{IQL} as each agent is trained without direct input from the other agents in the coalition. CTDE methods incorporate a centralised critic into their architecture at training time which is trained to perform multi-agent credit assignment and generates more informative reward signals for the agent in the coalition \citep{VDN,rashid2018qmix,yu2021}. This centralised critic is only used during training time. During execution time agents in the coalition act in an ad-hoc or decentralised manner like in the DTDE case.

\textbf{Value Decomposition. } In multi-agent systems, typically a global shared reward signal is generated. Training a MARL system from this signal is difficult as we cannot determine the contribution of each member of the coalition from the environment feedback which makes it hard to evaluate how effective each member's learnt policy is. Value decomposition network (VDN) \citep{VDN} introduced a new paradigm where a centralised critic is used to split the joint signal into individual agent rewards for better credit assignment. This method has proved very effective for Q-Learning based algorithms and became the most common avenue of development for new cooperative MARL methods \citep{rashid2018qmix,rashid2020weighted,mahajan2020maven,wang2020rode,wang2021qplex}. For policy gradient (PG) methods, multi-agent proximal policy optimisation (MAPPO) \citep{yu2021} has been shown to be highly effective in many cooperative settings. However, while it falls under the CTDE paradigm, it does not make use of value decomposition.

%% file: sections/3-Trends.tex
\section{Historical trends vs new data}

\begin{figure}[t]
    \centering
        \includegraphics[width=0.45\textwidth, valign=t]{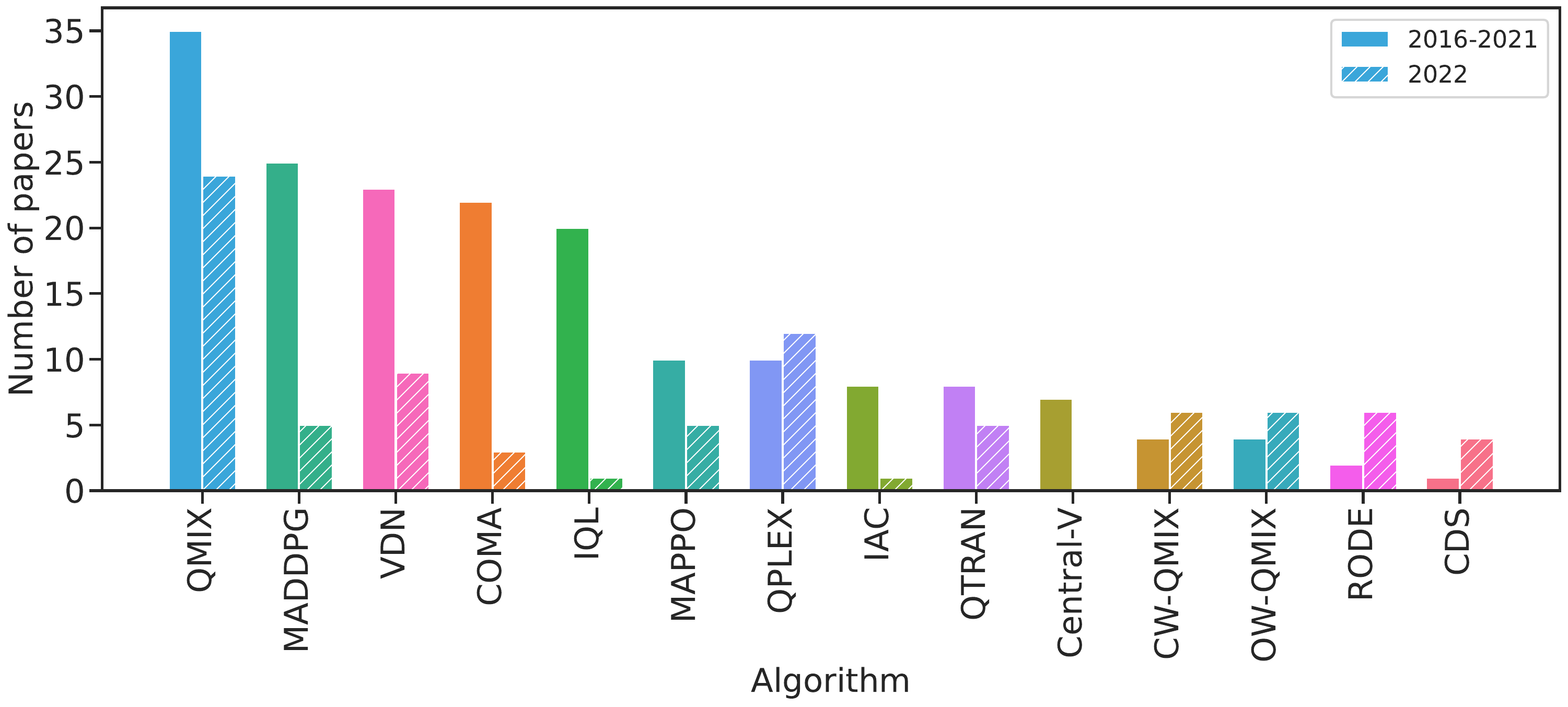}

    \caption{\textit{Comparison between occurrences of algorithms from past years against the most recent year of full data.}}
    \label{fig:algorithm_usage}
\end{figure}

\textbf{Depreciation of legacy algorithms. }The analysis of recent data indicates that, the landscape of common algorithms has shifted greatly from historical analysis as illustrated in figure \ref{fig:algorithm_usage}. Historic algorithms like COMA \citep{foerster2018counterfactual} and MADDPG \citep{lowe2017multi} have lost popularity as baselines with none of the reviewed papers making use of COMA and MADDPG used in only $25.8{\%}$ of new publications vs a historical use of $35.2{\%}$. As these methods are unable to achieve reasonable performance on current common benchmarks their decline reflects researchers moving towards more modern baselines \citep{papoudakis2021benchmarking}. Qmix \citep{rashid2018qmix} remains a strong  baseline for newer cooperative MARL algorithms which take inspiration from it. Importantly, Qmix is still competitive with newer methods when parameterised correctly and maintains relevance on modern benchmarks \citep{Hu-2021}. 

Despite the discovery of the effectiveness of Proximal Policy Optimisation (PPO)  \citep{schulman2017proximal} to match the more complex value decomposition methods in performance and sample efficiency \citep{yu2021}, PG approaches still seem unpopular. With both the CTDE and DTDE variants of the Advantage Actor-Critic (A2C) \citep{mnih2016asynchronous}; Independent Actor-Critic (IAC) and Central-V \citep{foerster2018counterfactual} not being present in newer publications. However MAPPO was still used in $16.1{\%}$ of the new papers and will hopefully gain traction as a common baseline. Another notable development is the decline of Independent Learners (IL). Although normally used as baselines, the findings in \citep{papoudakis2021benchmarking} show that they are important to assess the trade-offs newer method may have in certain settings.

\begin{figure}[t]
    \centering
    \begin{subfigure}[t]{0.02\textwidth}
        \scriptsize
        \textbf{(a)}
    \end{subfigure}
    \begin{subfigure}[t]{0.4\textwidth}
        \includegraphics[width=\textwidth, valign=t]{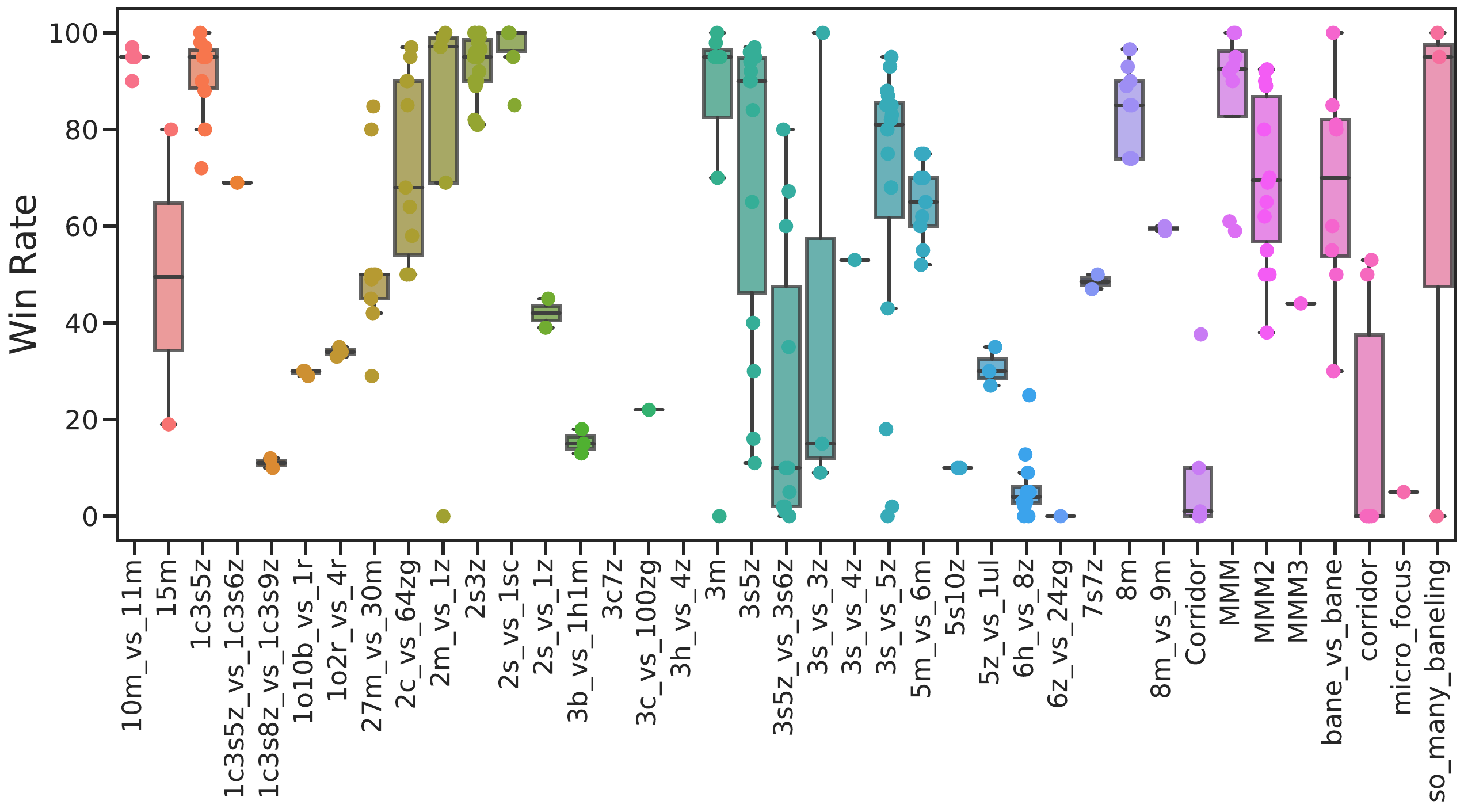}
        \captionlistentry{}
        \label{fig:perfomance_2022}
    \end{subfigure}
    
    \begin{subfigure}[t]{0.02\textwidth}
        \scriptsize
        \textbf{(b)}
    \end{subfigure}
    \begin{subfigure}[t]{0.4\textwidth}
        \includegraphics[width=\textwidth, valign=t]{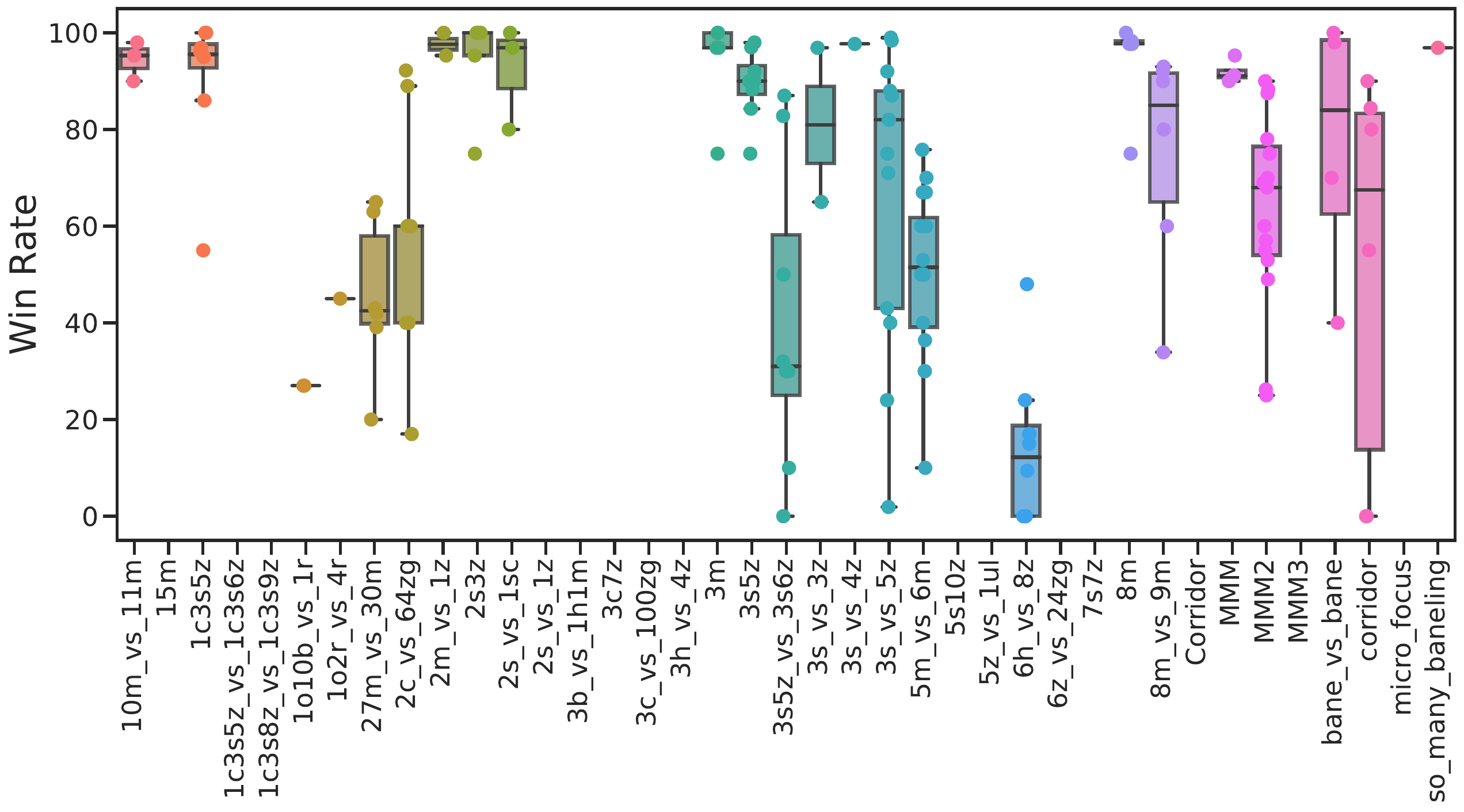}
        \captionlistentry{}
        \label{fig:perfomance_2023}
    \end{subfigure}
    \caption{\textit{Comparison between the performance of the Qmix algorithm from 2016-2021 against 2022.} \textbf{(a)} Historic performance spread for Qmix in SMAC (2016-2021). \textbf{(b)} Recent performance spread for Qmix in SMAC (2022).}
    \label{fig:performance}
\end{figure}

\begin{figure}[t]
    \centering
        \includegraphics[width=0.47\textwidth]{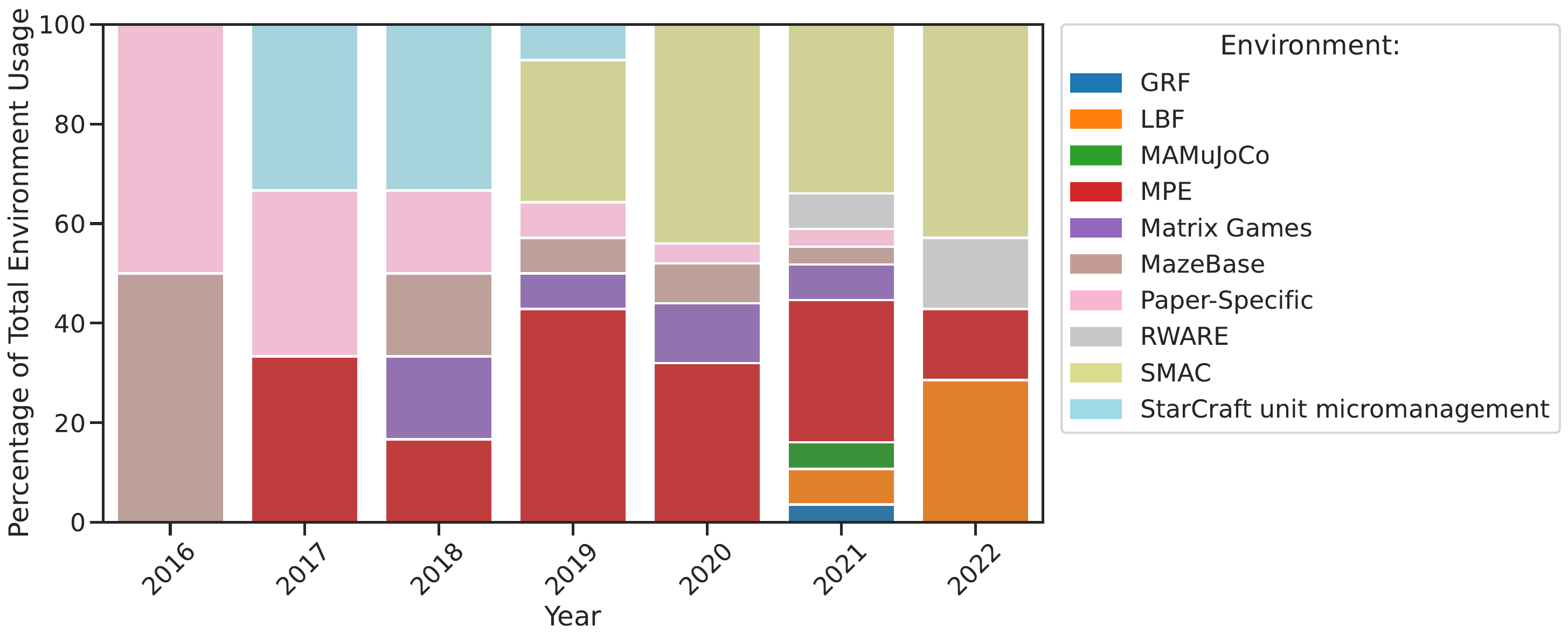}
        \caption{Historic environment usage from 2016 to 2022}
        \label{fig:environments}
\end{figure}

\textbf{Inconsistencies in performance for well-studied settings. }In the early years of MARL, there was naturally a high variance in results even when using the same algorithm. Initially standardised frameworks were uncommon and MARL algorithms suffered from high implementation variance. Fortunately, MARL experienced some standardisation with the release of the the Starcraft Multi-Agent Challenge (SMAC) \citep{SMAC} along with the EPyMARL framework \citep{papoudakis2021benchmarking}. Since then it has become and remains the most common benchmark in cooperative MARL by a sizable margin as shown in figure \ref{fig:environments}. However, even with this framework and common testing environment there was still high variance in reported performance as shown in figure \ref{fig:perfomance_2022}. For the most recent data in Figure \ref{fig:perfomance_2023} we can see that these trends persist. Seemingly, variance has only reduced significantly on the `easy` category of scenarios like 8m which are trivial for newer algorithms. In practice this makes it very difficult to determine the true performance gains of recently developed and future works against the historical baselines they aim to improve on.\looseness=-1 

\begin{figure}[t]
    \centering
    \begin{subfigure}[t]{0.02\textwidth}
        \scriptsize
        \textbf{(a)}
    \end{subfigure}
    \begin{subfigure}[t]{0.4\textwidth}
        \includegraphics[width=\textwidth, valign=t]{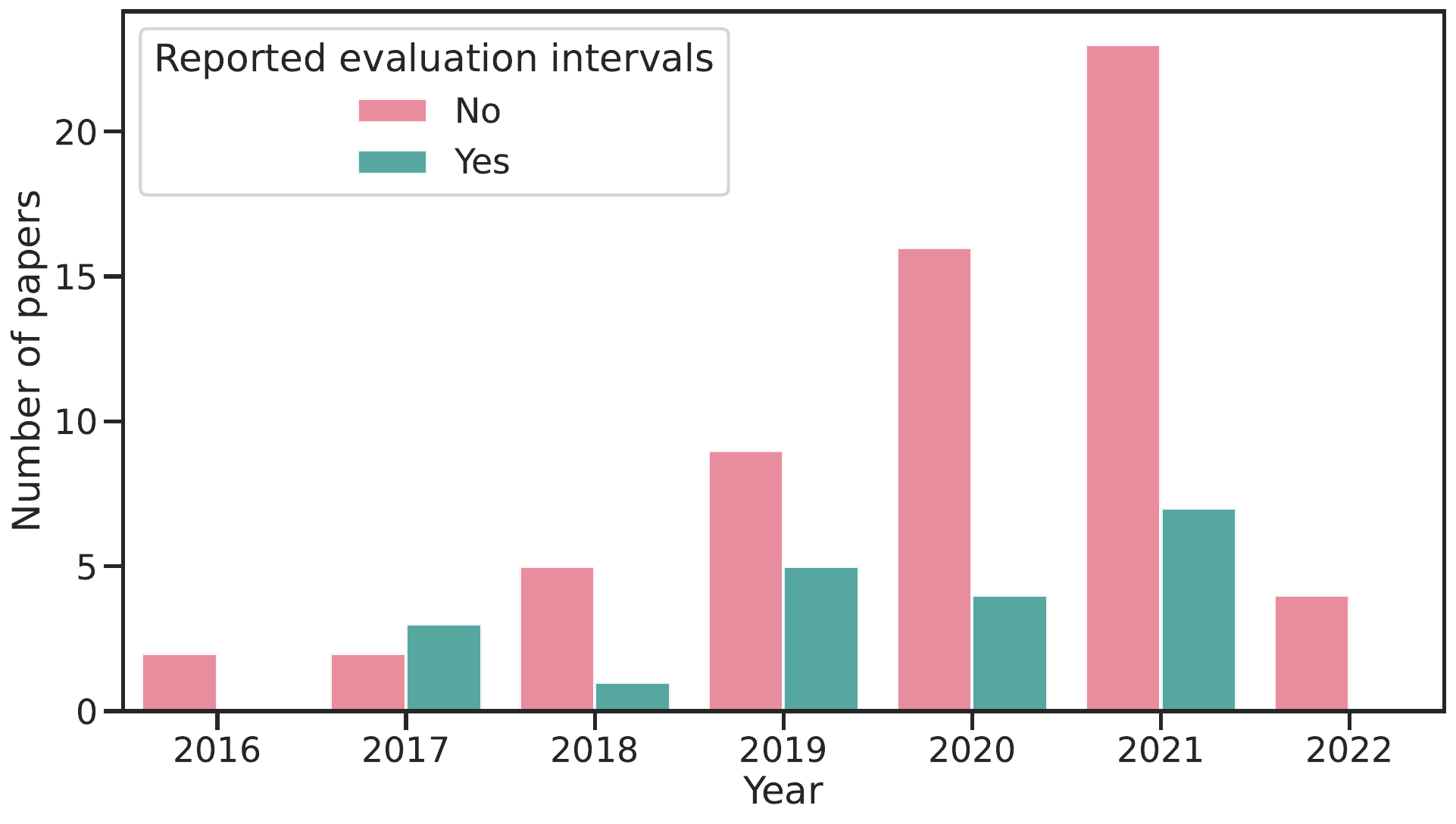}
        \captionlistentry{}
        \label{fig: evaluation_reporting}
    \end{subfigure}
    
    \begin{subfigure}[t]{0.02\textwidth}
        \scriptsize
        \textbf{(b)}
    \end{subfigure}
    \begin{subfigure}[t]{0.4\textwidth}
        \includegraphics[width=\textwidth, valign=t]{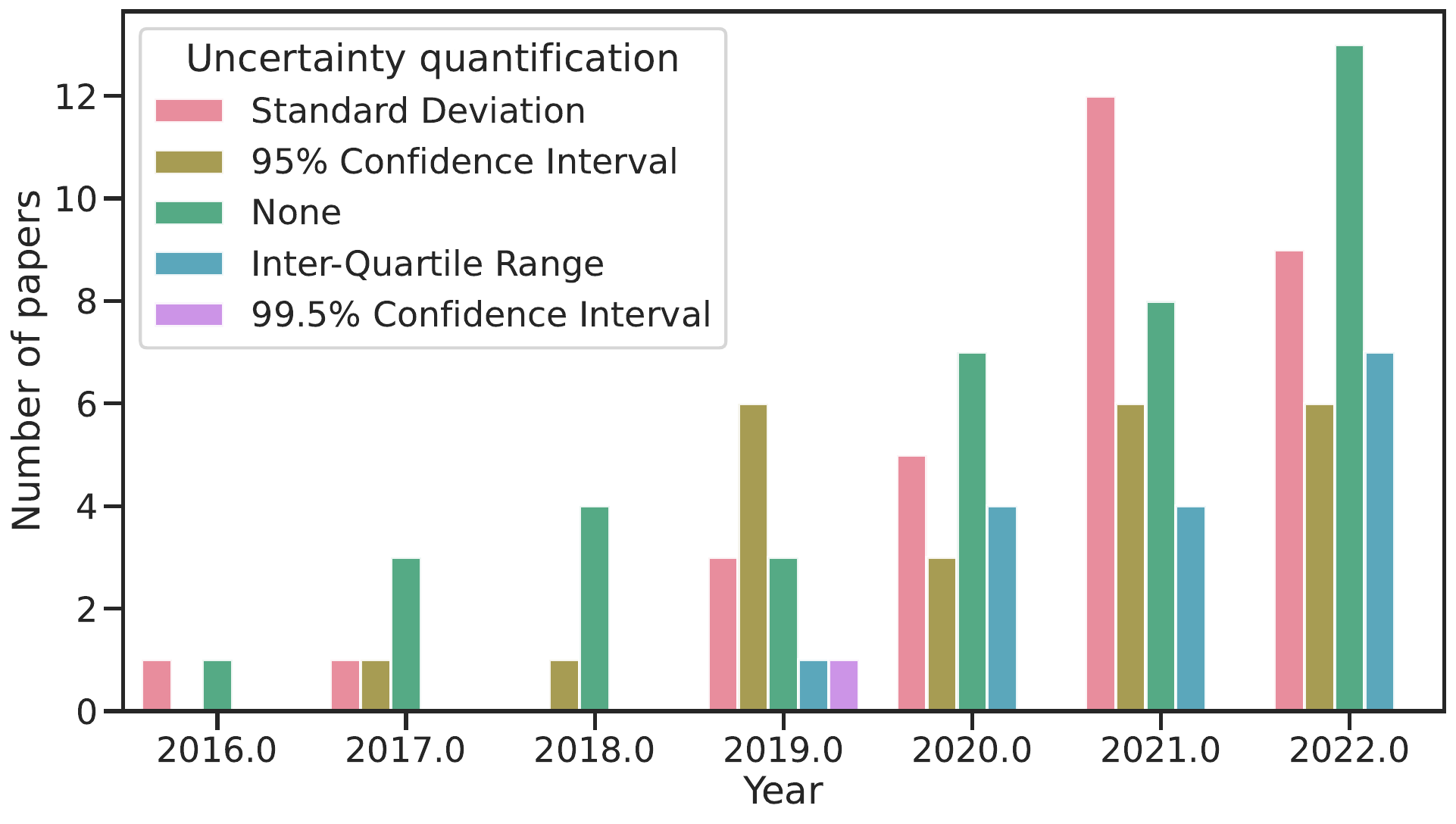}
        \captionlistentry{}
        \label{fig:uncertainty_2023}
    \end{subfigure}
    \caption{\textit{Historical tendencies for reporting evaluation runs and uncertainty metrics from 2016 to 2022} \textbf{(a)} Tendencies for reporting evaluation runs (2016-2022). \textbf{(b)} Tendencies for reporting uncertainty metrics (2016-2022).}
    \label{fig:eval_uncertainty}
\end{figure}

\textbf{Tends in performance reporting. }Metric reporting reached a promising peak of $75{\%}$ in 2021 and seemed to be a rising trend over previous years. However we see this drop to $63{\%}$ in 2022 in figure \ref{fig: evaluation_reporting}. This is alarming given the goal of using MARL in real-world settings where reliability is often more important than absolute performance. 

\textbf{Performance Aggregation. }Additionally, aggregation metrics are not present in $20{\%}$ of recent publications and with the computational complexity of modern MARL it is very difficult to evaluate over enough seeds to account for the variance in performance over different runs \citep{agarwal2022deep}.\looseness=-1 

\begin{figure}[h]
    \centering
    \begin{subfigure}[t]{0.02\textwidth}
        \scriptsize
        \textbf{(a)}
    \end{subfigure}
    \begin{subfigure}[t]{0.18\textwidth}
        \includegraphics[width=\textwidth, valign=t]{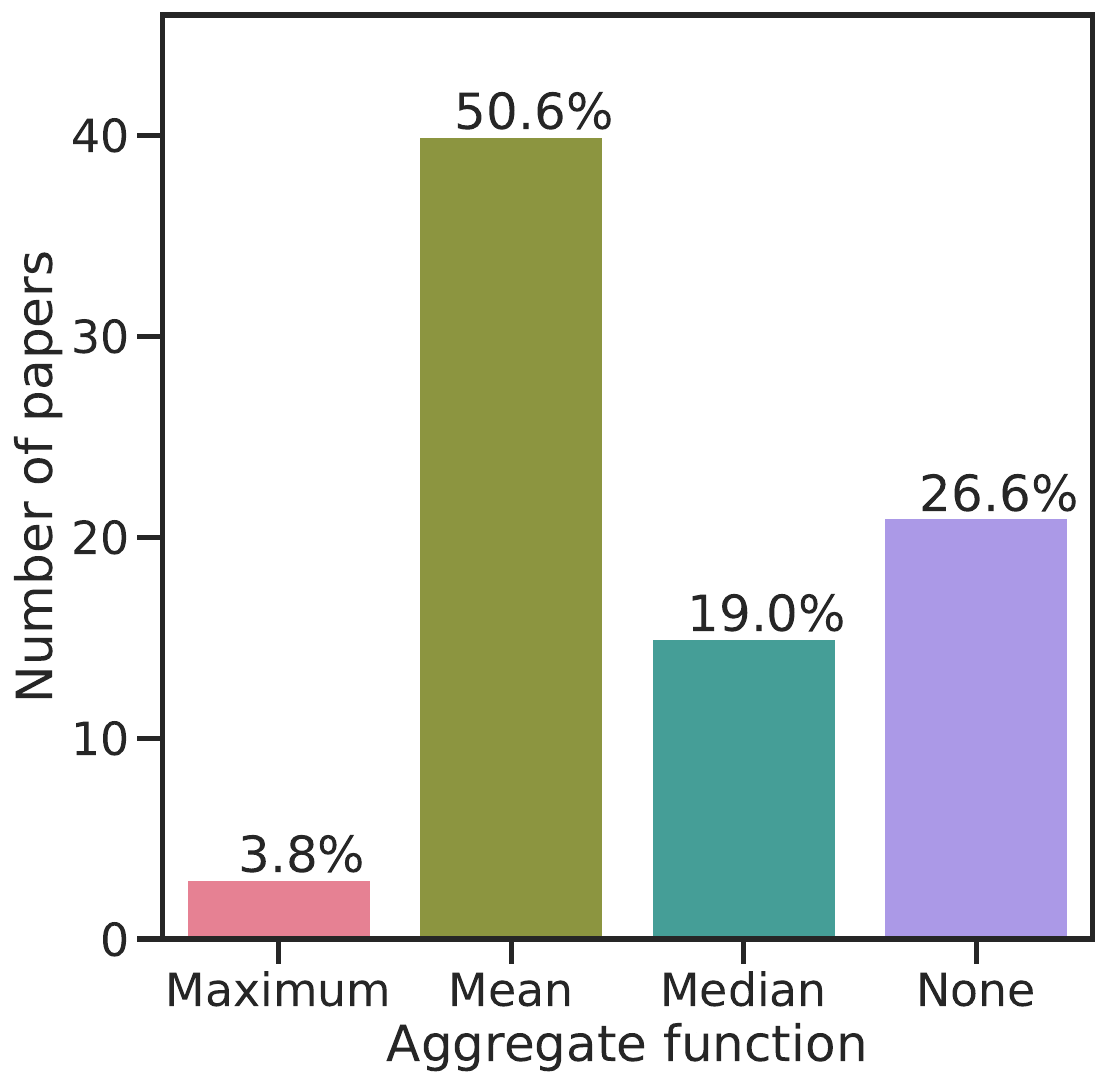}
        \captionlistentry{}
        \label{fig:aggregation_2022}
    \end{subfigure}
    \begin{subfigure}[t]{0.02\textwidth}
        \scriptsize
        \textbf{(b)}
    \end{subfigure}
    \begin{subfigure}[t]{0.18\textwidth}
        \includegraphics[width=\textwidth, valign=t]{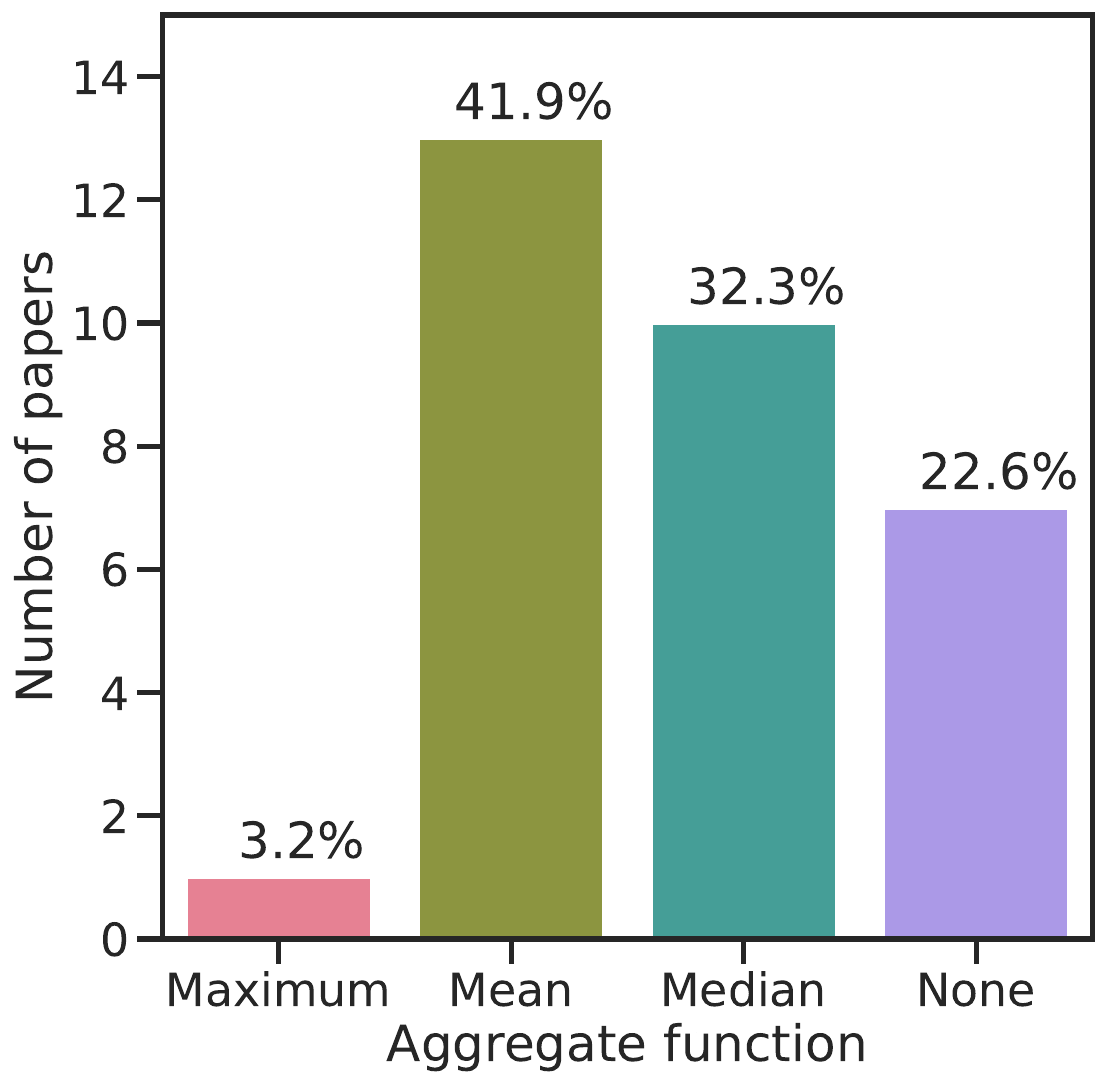}
        \captionlistentry{}
        \label{aggregation_2023}
    \end{subfigure}
    
    \begin{subfigure}[t]{0.02\textwidth}
        \scriptsize
        \textbf{(c)}
    \end{subfigure}
    \begin{subfigure}[t]{0.18\textwidth}
        \includegraphics[width=\textwidth, valign=t]{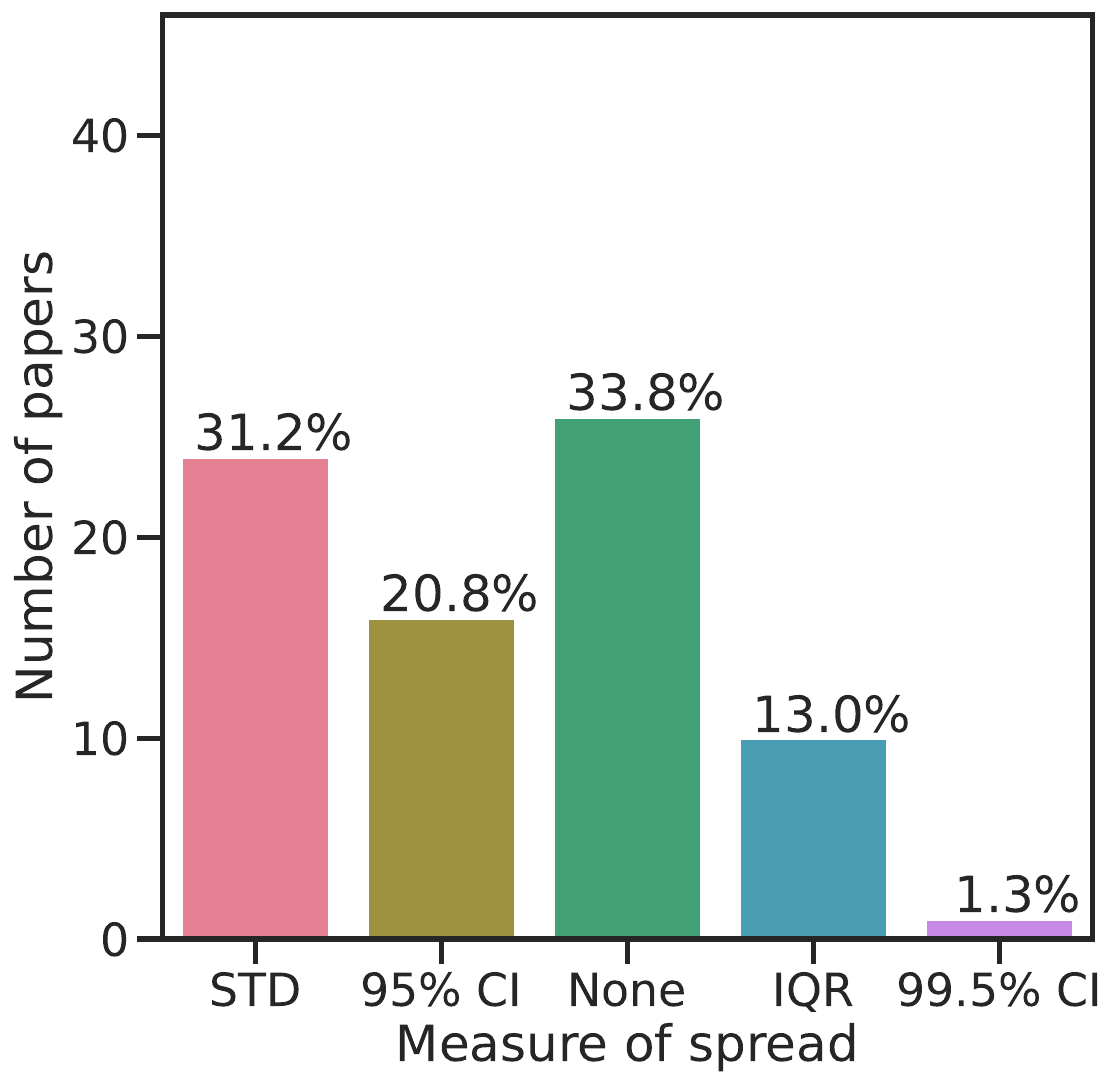}
        \captionlistentry{}
        \label{fig:spread_2022}
    \end{subfigure}
    \begin{subfigure}[t]{0.02\textwidth}
        \scriptsize
        \textbf{(d)}
    \end{subfigure}
    \begin{subfigure}[t]{0.18\textwidth}
        \includegraphics[width=\textwidth, valign=t]{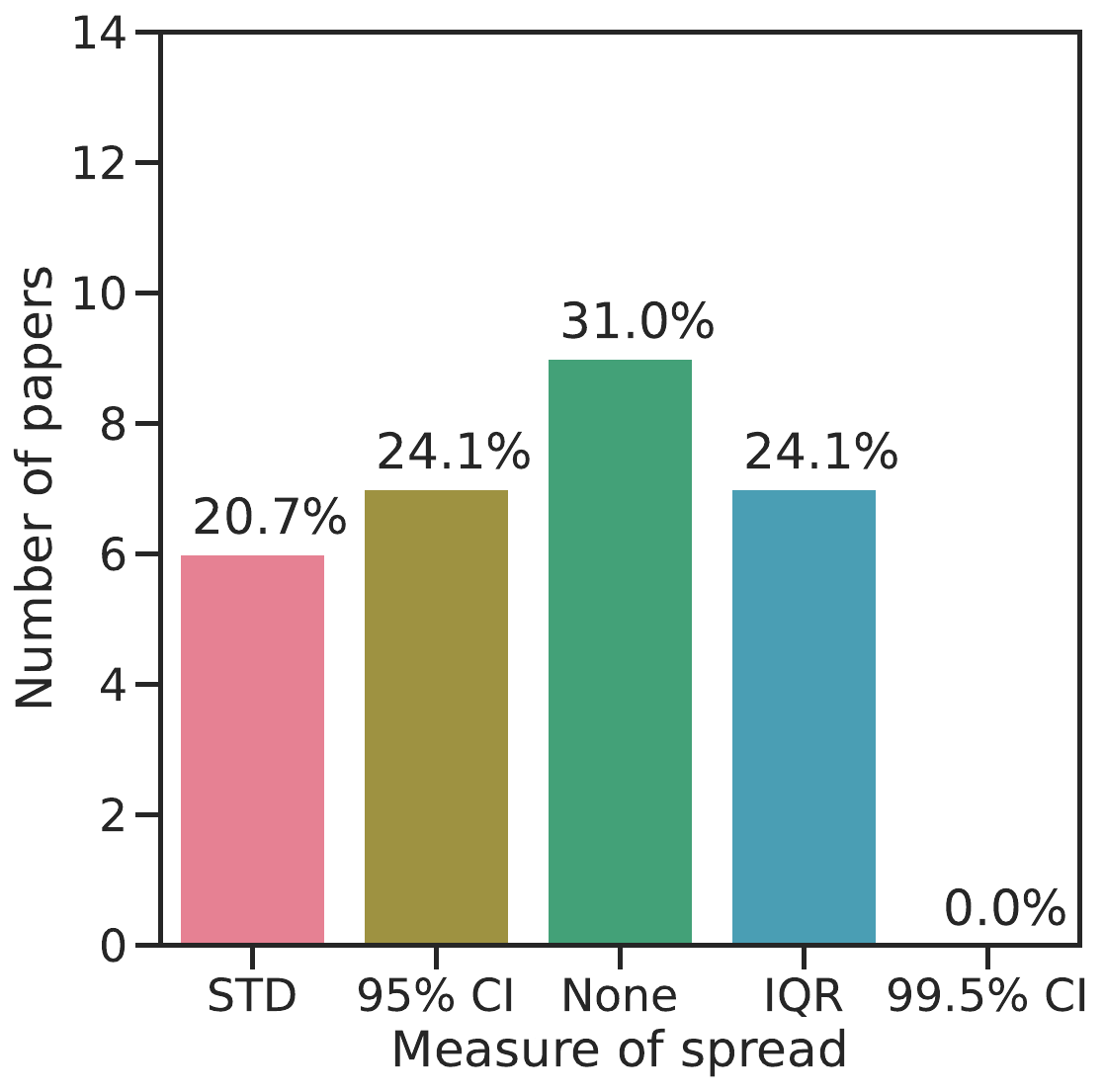}
        \captionlistentry{}
        \label{fig:spread_2023}
    \end{subfigure}
    \caption{\textit{Historical tendencies for reporting performance aggregation and variance from 2016 to 2022} \textbf{(a)} Historical aggregation usage (2016-2021). \textbf{(b)} Recent aggregation usage (2022). \textbf{(c)} Historical measure of spread (2016-2021). \textbf{(d)} Recent measure of spread (2022).}
    \label{fig:aggregation}
\end{figure}

\textbf{Measurements of spread and uncertainty. }From figures \ref{fig:spread_2022} and \ref{fig:spread_2023}, it is clear that the tendency to display a measurement of spread has not meaningfully changed going from 33.8\% to 31\% respectively. Similarly, from figure \ref{fig:uncertainty_2023} it is clear that the number of papers not reporting uncertainty are also increasing relative to the total number of published papers each year. This statistic is alarming most deep RL publications rarely use enough environment seeds to account for algorithmic variance during training \citep{agarwal2022deep}. Further, for many practical tasks where high safety is a priority, methods with highly inconsistent behaviors are not usable which hampers the development of algorithms that have real-world utility.

\textbf{Environment usage trends. }As mentioned previously, SMAC is the most popular environment in MARL by a large margin followed by MPE which can be seen in figure \ref{fig:environments}. Evidence suggests that these settings have both reached a point of being over-fit to \citep{papoudakis2021benchmarking, gorsane2022standardised} which makes determining algorithmic ranking difficult. Newer settings have generally not gained much popularity and despite the insight RWARE and LBF were shown to provide on MARL algorithms by  \citep{papoudakis2021benchmarking} recent data shows they are underutilised. However, environment overfitting on SMAC may be naturally resolved as users move over to SMAC-v2 \citep{ellis2022smacv2} which provides a new host of challenges through a well-understood interface. Given these findings it may be more effective to make use of explainability frameworks like ShinRL \citep{ShinRL} to determine a human interpreble  representation of the learnt policy and from this compare the viability and variability of different methods. Even without a fully-featured framework, using methods like those described in \citep{agogino2008analyzing,albrecht2019reasoning,Boggess2022TowardPE} we can uncover exactly what different settings are testing and the limitations of current MARL methods that are difficult to detect from pure performance plots.

\begin{figure}[h]
    \centering
    \begin{subfigure}[t]{0.15\textwidth}
        \includegraphics[width=\textwidth, valign=t]{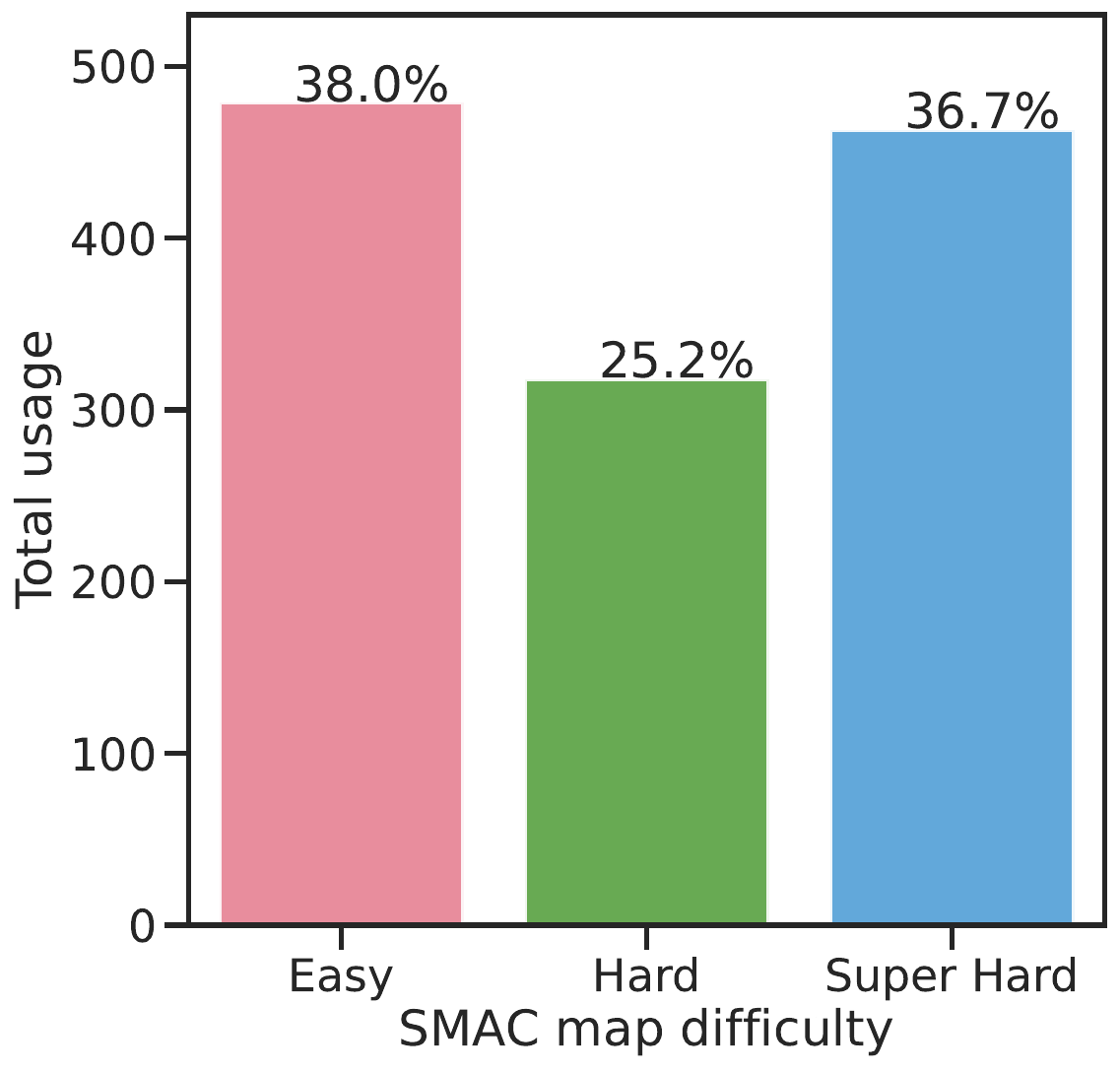}
        \captionlistentry{}
        \label{fig:smac_2022}
    \end{subfigure}
    \begin{subfigure}[t]{0.15\textwidth}
        \includegraphics[width=\textwidth, valign=t]{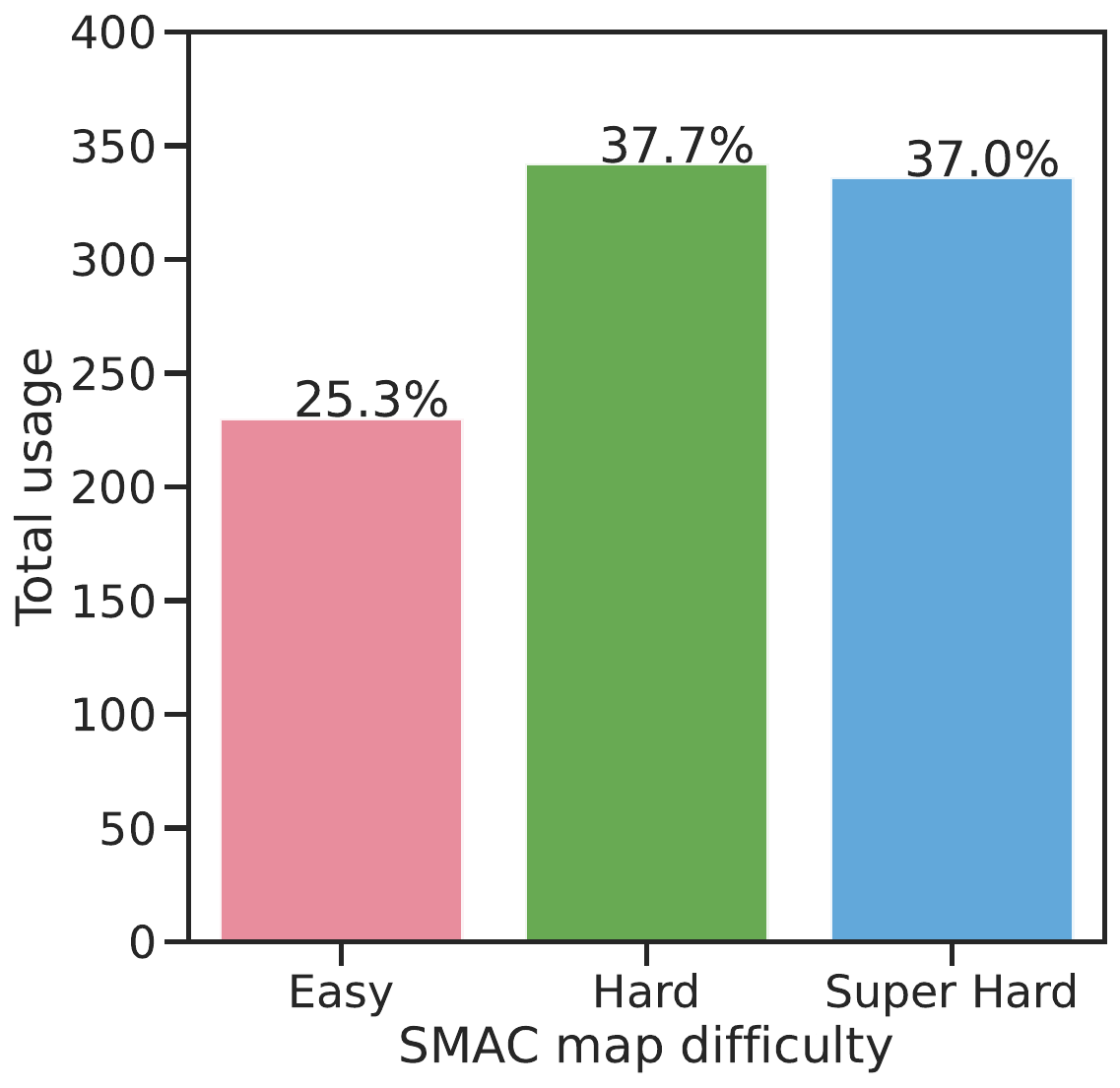}
        \captionlistentry{}
        \label{smac_2023}
    \end{subfigure}
     \begin{subfigure}[t]{0.15\textwidth}
        \includegraphics[width=\textwidth, valign=t]{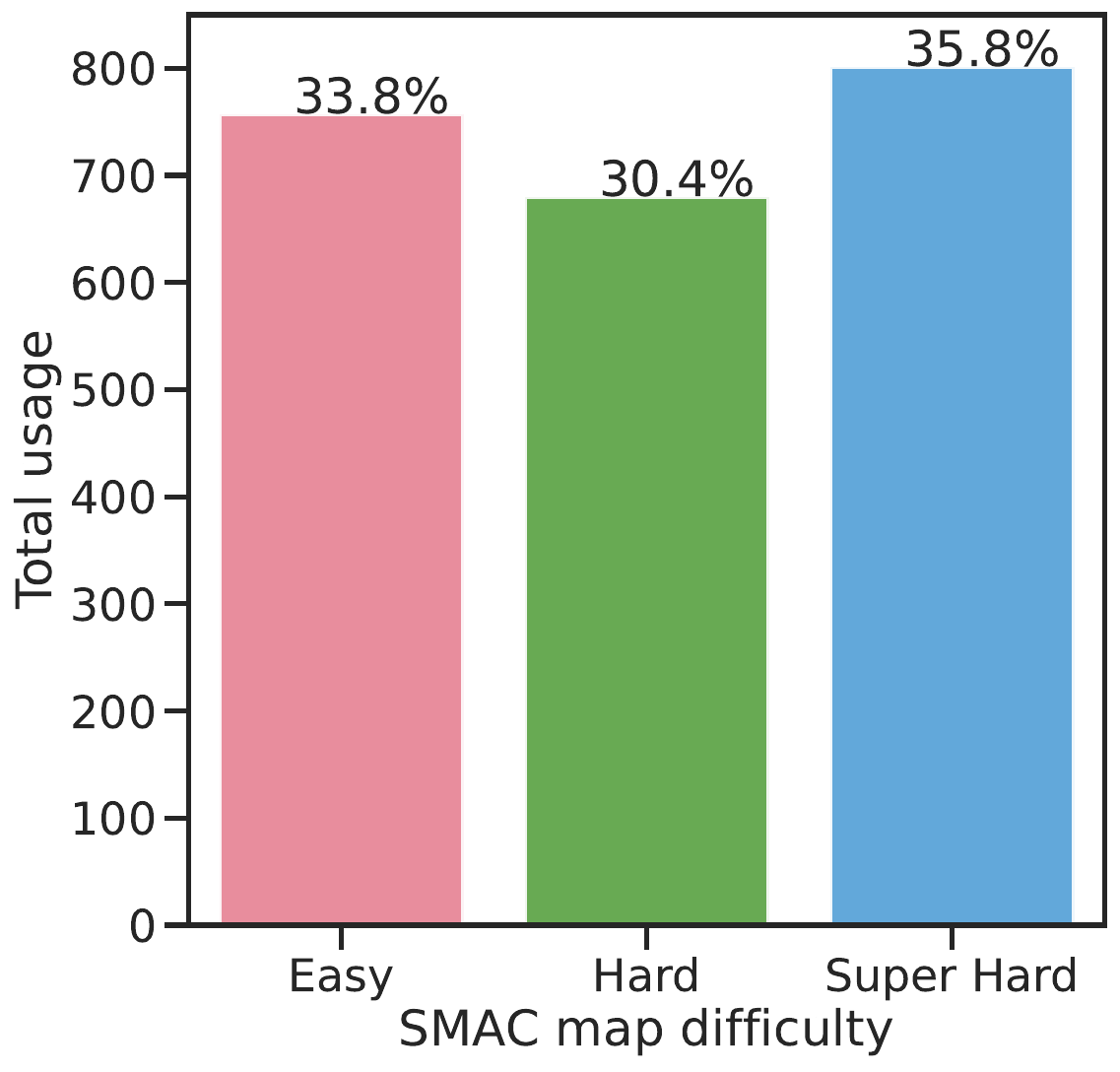}
        \captionlistentry{}
        \label{smac_all}
    \end{subfigure}
    \caption{\textit{Historical SMAC environment difficulty usage since release} \textbf{(Left)} Historical SMAC scenario usage (2018-2021). \textbf{(Middle)} Recent SMAC scenario usage (2022). \textbf{(Right)} All SMAC scenario usage (2018-2022).}
    \label{fig:smac_scenarios}
\end{figure}

\textbf{SMAC scenario usage. }SMAC categorises scenarios into `easy`, `hard` and `super hard` difficulties. Over time, scenarios not in the `super hard` category have become trivial to solve. Due to the large number of scenarios in SMAC, authors will often use a subset of the scenarios to reduce compute time which produces misleading aggregate performance \citep{gorsane2022standardised}. A helpful development seen in figure \ref{fig:smac_scenarios} is the reduction in use of the `easy` scenarios as a focus on difficult challenges streamlines the total set of evaluation scenarios required for comparative evaluation. \citep{gorsane2022standardised} advocate for environment creators to propose a minimal viable set of scenarios for their settings along with an evaluation guideline to improve evaluation, however, when done by hand this induces human bias and until a setting has been overfit to it can be difficult for authors to determine the relevancy of each scenario. By using methods that explain environment features like \citep{hu2022policy} we can disentangle what competencies are required by each scenario and what exactly what is needed for the algorithm to solve the setting.

%% file: sections/4-solutions.tex
\section{Recent positive developments}

\textbf{On environment overfitting and computational expense. } Along with the signs of overfitting, the authors of SMAC-v1 have done an analysis of their environment and discovered that due to its deterministic nature, it can be solved purely by memorising timesteps \citep{ellis2022smacv2}. They have noted that algorithms designed for this type of setting at not suitable for the stochastic nature of real-world problems and have noted how in the single-agent setting procedural  content generation (pcg) has gained popularity in resolving this issue \citep{juliani2019obstacle, küttler2020nethack, cobbe2020leveraging, Samvelyan2021MiniHackTP, MinigridMiniworld23}. However, although SMAC-v2 does resolve the limitations of SMAC-v1 w.r.t. generalisation and overfitting, it can still be computationally expensive to conform to the required number of runs outlined in works like \citep{agarwal2022deep,gorsane2022standardised} as it relies on a full-fledged video game as a back end. Fortunately there have been some promising developments in using JAX \citep{jax2018github} for RL applications. Most notably JAX-based RL settings \citep{brax2021github,gymnax2022github,bonnet2023jumanji}, which can be run on computational-accelerators like TPUs and GPUs. This allows the entire RL training loop to be massively accelerated. 
Growth of these types of settings have great potential to increase the clarity of RL by making experiments replicable for researchers without access to massive amounts of compute and allow more democratised scrutiny of RL research as a whole. 

\textbf{Emerging XAI in MARL. } Although there has been an increase in XAI literature for SARL \citep{juozapaitis2019explainable, madumal2020explainable, puiutta2020explainable, glanois2021survey, heuillet2021explainability, vouros2022explainable,dazeley2023explainable} and there are now many methods, frameworks and environment suites to make RL human-interpretable, MARL has been more historically limited. However, recent works have attempted to use XAI to explain not only agent behaviors but also what features are present in different settings. Notably \citep{hu2022policy} make use of algorithm-agnostic metrics to determine which common implementation features in MARL effect the performance of algorithms in popular settings. Further development of methods like this will greatly improve the clarity of MARL algorithms and improve trust in the field when attempting to apply these methods to real-world settings. Additionally, they are able to provide us with clearer paths for which areas in MARL require attention when compared to raw performance metrics.

%% file: sections/5-Conclusion.tex
\section{Conclusion}

We compiled 29 papers from the latest MARL publications to extend the MARL publication database to take into account recent development trends.

Despite a push for increased rigour, certain worrying historical trends are still prevalent in MARL w.r.t. replicability. Qmix, despite it's longstanding use as a baseline, still shows large variance across publications. The details required to replicate evaluation methodologies are often unreported and IL baselines are absent. Together, these issues make it difficult to get a true measurement of algorithmic development against historic algorithms.

Additionally environment usage also seems to have not yet diversified and research is focused on SMAC and MPE. However, authors do seem to be moving towards a more streamlined set of scenarios which will reduce computational overhead. Given these trends it is important that replicability is approached proactively by the MARL community to ensure trust in the field is maintained as it grows.

%% file: main.bbl
\begin{thebibliography}{56}
\providecommand{\natexlab}[1]{#1}

\bibitem[{Agarwal et~al.(2022)Agarwal, Schwarzer, Castro, Courville, and Bellemare}]{agarwal2022deep}
Agarwal, R.; Schwarzer, M.; Castro, P.~S.; Courville, A.; and Bellemare, M.~G. 2022.
\newblock {Deep Reinforcement Learning at the Edge of the Statistical Precipice}.
\newblock arXiv:2108.13264.

\bibitem[{Agogino and Tumer(2008)}]{agogino2008analyzing}
Agogino, A.~K.; and Tumer, K. 2008.
\newblock Analyzing and visualizing multiagent rewards in dynamic and stochastic domains.
\newblock \emph{Autonomous Agents and Multi-Agent Systems}, 17: 320--338.

\bibitem[{Albrecht and Stone(2019)}]{albrecht2019reasoning}
Albrecht, S.~V.; and Stone, P. 2019.
\newblock Reasoning about Hypothetical Agent Behaviours and their Parameters.
\newblock arXiv:1906.11064.

\bibitem[{Bakhtin et~al.(2022)Bakhtin, Brown, Dinan, Farina, Flaherty, Fried, Goff, Gray, Hu, Jacob, Komeili, Konath, Kwon, Lerer, Lewis, Miller, Mitts, Renduchintala, Roller, Rowe, Shi, Spisak, Wei, Wu, Zhang, and Zijlstra}]{Bakhtin2022HumanlevelPI}
Bakhtin, A.; Brown, N.; Dinan, E.; Farina, G.; Flaherty, C.; Fried, D.; Goff, A.; Gray, J.; Hu, H.; Jacob, A.~P.; Komeili, M.; Konath, K.; Kwon, M.; Lerer, A.; Lewis, M.; Miller, A.~H.; Mitts, S.; Renduchintala, A.; Roller, S.; Rowe, D.; Shi, W.; Spisak, J.; Wei, A.; Wu, D.~J.; Zhang, H.; and Zijlstra, M. 2022.
\newblock Human-level play in the game of Diplomacy by combining language models with strategic reasoning.
\newblock \emph{Science}, 378: 1067 -- 1074.

\bibitem[{Boggess, Kraus, and Feng(2022)}]{Boggess2022TowardPE}
Boggess, K.; Kraus, S.; and Feng, L. 2022.
\newblock Toward Policy Explanations for Multi-Agent Reinforcement Learning.
\newblock In \emph{International Joint Conference on Artificial Intelligence}.

\bibitem[{Bonnet et~al.(2023)Bonnet, Luo, Byrne, Surana, Coyette, Duckworth, Midgley, Kalloniatis, Abramowitz, Waters, Smit, Grinsztajn, Sob, Mahjoub, Tegegn, Mimouni, Boige, de~Kock, Furelos-Blanco, Le, Pretorius, and Laterre}]{bonnet2023jumanji}
Bonnet, C.; Luo, D.; Byrne, D.; Surana, S.; Coyette, V.; Duckworth, P.; Midgley, L.~I.; Kalloniatis, T.; Abramowitz, S.; Waters, C.~N.; Smit, A.~P.; Grinsztajn, N.; Sob, U. A.~M.; Mahjoub, O.; Tegegn, E.; Mimouni, M.~A.; Boige, R.; de~Kock, R.; Furelos-Blanco, D.; Le, V.; Pretorius, A.; and Laterre, A. 2023.
\newblock Jumanji: a Diverse Suite of Scalable Reinforcement Learning Environments in JAX.
\newblock arXiv:2306.09884.

\bibitem[{Bradbury et~al.(2018)Bradbury, Frostig, Hawkins, Johnson, Leary, Maclaurin, Necula, Paszke, Vander{P}las, Wanderman-{M}ilne, and Zhang}]{jax2018github}
Bradbury, J.; Frostig, R.; Hawkins, P.; Johnson, M.~J.; Leary, C.; Maclaurin, D.; Necula, G.; Paszke, A.; Vander{P}las, J.; Wanderman-{M}ilne, S.; and Zhang, Q. 2018.
\newblock {JAX}: composable transformations of {P}ython+{N}um{P}y programs.

\bibitem[{Brittain and Wei(2019)}]{brittain2019autonomous}
Brittain, M.; and Wei, P. 2019.
\newblock Autonomous Air Traffic Controller: A Deep Multi-Agent Reinforcement Learning Approach.
\newblock arXiv:1905.01303.

\bibitem[{Chevalier-Boisvert et~al.(2023)Chevalier-Boisvert, Dai, Towers, de~Lazcano, Willems, Lahlou, Pal, Castro, and Terry}]{MinigridMiniworld23}
Chevalier-Boisvert, M.; Dai, B.; Towers, M.; de~Lazcano, R.; Willems, L.; Lahlou, S.; Pal, S.; Castro, P.~S.; and Terry, J. 2023.
\newblock Minigrid \& Miniworld: Modular \& Customizable Reinforcement Learning Environments for Goal-Oriented Tasks.
\newblock \emph{CoRR}, abs/2306.13831.

\bibitem[{Cobbe et~al.(2020)Cobbe, Hesse, Hilton, and Schulman}]{cobbe2020leveraging}
Cobbe, K.; Hesse, C.; Hilton, J.; and Schulman, J. 2020.
\newblock Leveraging Procedural Generation to Benchmark Reinforcement Learning.
\newblock arXiv:1912.01588.

\bibitem[{Colas, Sigaud, and Oudeyer(2018)}]{colas2018random}
Colas, C.; Sigaud, O.; and Oudeyer, P.-Y. 2018.
\newblock {How Many Random Seeds? Statistical Power Analysis in Deep Reinforcement Learning Experiments}.
\newblock arXiv:1806.08295.

\bibitem[{Colas, Sigaud, and Oudeyer(2019)}]{colas2019hitchhikers}
Colas, C.; Sigaud, O.; and Oudeyer, P.-Y. 2019.
\newblock {A Hitchhiker's Guide to Statistical Comparisons of Reinforcement Learning Algorithms}.
\newblock arXiv:1904.06979.

\bibitem[{Dazeley, Vamplew, and Cruz(2023)}]{dazeley2023explainable}
Dazeley, R.; Vamplew, P.; and Cruz, F. 2023.
\newblock Explainable reinforcement learning for broad-xai: a conceptual framework and survey.
\newblock \emph{Neural Computing and Applications}, 1--24.

\bibitem[{Ellis et~al.(2022)Ellis, Moalla, Samvelyan, Sun, Mahajan, Foerster, and Whiteson}]{ellis2022smacv2}
Ellis, B.; Moalla, S.; Samvelyan, M.; Sun, M.; Mahajan, A.; Foerster, J.~N.; and Whiteson, S. 2022.
\newblock SMACv2: An Improved Benchmark for Cooperative Multi-Agent Reinforcement Learning.

\bibitem[{Engstrom et~al.(2020)Engstrom, Ilyas, Santurkar, Tsipras, Janoos, Rudolph, and Madry}]{engstrom2020implementation}
Engstrom, L.; Ilyas, A.; Santurkar, S.; Tsipras, D.; Janoos, F.; Rudolph, L.; and Madry, A. 2020.
\newblock Implementation Matters in Deep Policy Gradients: A Case Study on PPO and TRPO.
\newblock arXiv:2005.12729.

\bibitem[{Foerster et~al.(2018)Foerster, Farquhar, Afouras, Nardelli, and Whiteson}]{foerster2018counterfactual}
Foerster, J.; Farquhar, G.; Afouras, T.; Nardelli, N.; and Whiteson, S. 2018.
\newblock Counterfactual multi-agent policy gradients.
\newblock In \emph{Proceedings of the AAAI conference on artificial intelligence}, volume~32.

\bibitem[{Foerster et~al.(2019)Foerster, Song, Hughes, Burch, Dunning, Whiteson, Botvinick, and Bowling}]{foerster2019bayesian}
Foerster, J.~N.; Song, F.; Hughes, E.; Burch, N.; Dunning, I.; Whiteson, S.; Botvinick, M.; and Bowling, M. 2019.
\newblock Bayesian Action Decoder for Deep Multi-Agent Reinforcement Learning.
\newblock arXiv:1811.01458.

\bibitem[{Freeman et~al.(2021)Freeman, Frey, Raichuk, Girgin, Mordatch, and Bachem}]{brax2021github}
Freeman, C.~D.; Frey, E.; Raichuk, A.; Girgin, S.; Mordatch, I.; and Bachem, O. 2021.
\newblock Brax - A Differentiable Physics Engine for Large Scale Rigid Body Simulation.

\bibitem[{Glanois et~al.(2021)Glanois, Weng, Zimmer, Li, Yang, Hao, and Liu}]{glanois2021survey}
Glanois, C.; Weng, P.; Zimmer, M.; Li, D.; Yang, T.; Hao, J.; and Liu, W. 2021.
\newblock A survey on interpretable reinforcement learning.
\newblock \emph{arXiv preprint arXiv:2112.13112}.

\bibitem[{Gorsane et~al.(2022)Gorsane, Mahjoub, de~Kock, Dubb, Singh, and Pretorius}]{gorsane2022standardised}
Gorsane, R.; Mahjoub, O.; de~Kock, R.; Dubb, R.; Singh, S.; and Pretorius, A. 2022.
\newblock Towards a Standardised Performance Evaluation Protocol for Cooperative MARL.
\newblock arXiv:2209.10485.

\bibitem[{Gu et~al.(2021)Gu, Zhang, Lin, and Alazab}]{InternetofControllableThings}
Gu, B.; Zhang, X.; Lin, Z.; and Alazab, M. 2021.
\newblock Deep Multiagent Reinforcement-Learning-Based Resource Allocation for Internet of Controllable Things.
\newblock \emph{IEEE Internet of Things Journal}, 8(5): 3066--3074.

\bibitem[{Henderson et~al.(2018)Henderson, Islam, Bachman, Pineau, Precup, and Meger}]{henderson2019deep}
Henderson, P.; Islam, R.; Bachman, P.; Pineau, J.; Precup, D.; and Meger, D. 2018.
\newblock {Deep Reinforcement Learning that Matters}.
\newblock arXiv:1709.06560.

\bibitem[{Heuillet, Couthouis, and D{\'\i}az-Rodr{\'\i}guez(2021)}]{heuillet2021explainability}
Heuillet, A.; Couthouis, F.; and D{\'\i}az-Rodr{\'\i}guez, N. 2021.
\newblock Explainability in deep reinforcement learning.
\newblock \emph{Knowledge-Based Systems}, 214: 106685.

\bibitem[{Hu and Foerster(2021)}]{hu2021simplified}
Hu, H.; and Foerster, J.~N. 2021.
\newblock Simplified Action Decoder for Deep Multi-Agent Reinforcement Learning.
\newblock arXiv:1912.02288.

\bibitem[{Hu et~al.(2021)Hu, Jiang, Harding, Wu, and Liao}]{Hu-2021}
Hu, J.; Jiang, S.; Harding, S.~A.; Wu, H.; and Liao, S.-w. 2021.
\newblock Rethinking the Implementation Tricks and Monotonicity Constraint in Cooperative Multi-Agent Reinforcement Learning.

\bibitem[{Hu et~al.(2022)Hu, Xie, Liang, and Chang}]{hu2022policy}
Hu, S.; Xie, C.; Liang, X.; and Chang, X. 2022.
\newblock We have only.
\newblock arXiv:2207.05683.

\bibitem[{Jordan et~al.(2020)Jordan, Chandak, Cohen, Zhang, and Thomas}]{jordan2020evaluating}
Jordan, S.~M.; Chandak, Y.; Cohen, D.; Zhang, M.; and Thomas, P.~S. 2020.
\newblock {Evaluating the Performance of Reinforcement Learning Algorithms}.
\newblock arXiv:2006.16958.

\bibitem[{Juliani et~al.(2019)Juliani, Khalifa, Berges, Harper, Teng, Henry, Crespi, Togelius, and Lange}]{juliani2019obstacle}
Juliani, A.; Khalifa, A.; Berges, V.-P.; Harper, J.; Teng, E.; Henry, H.; Crespi, A.; Togelius, J.; and Lange, D. 2019.
\newblock Obstacle Tower: A Generalization Challenge in Vision, Control, and Planning.
\newblock arXiv:1902.01378.

\bibitem[{Juozapaitis et~al.(2019)Juozapaitis, Koul, Fern, Erwig, and Doshi-Velez}]{juozapaitis2019explainable}
Juozapaitis, Z.; Koul, A.; Fern, A.; Erwig, M.; and Doshi-Velez, F. 2019.
\newblock Explainable reinforcement learning via reward decomposition.
\newblock In \emph{IJCAI/ECAI Workshop on explainable artificial intelligence}.

\bibitem[{Kitamura and Yonetani(2021)}]{ShinRL}
Kitamura, T.; and Yonetani, R. 2021.
\newblock ShinRL: A Library for Evaluating RL Algorithms from Theoretical and Practical Perspectives.

\bibitem[{Küttler et~al.(2020)Küttler, Nardelli, Miller, Raileanu, Selvatici, Grefenstette, and Rocktäschel}]{küttler2020nethack}
Küttler, H.; Nardelli, N.; Miller, A.~H.; Raileanu, R.; Selvatici, M.; Grefenstette, E.; and Rocktäschel, T. 2020.
\newblock The NetHack Learning Environment.
\newblock arXiv:2006.13760.

\bibitem[{Lange(2022)}]{gymnax2022github}
Lange, R.~T. 2022.
\newblock {gymnax}: A {JAX}-based Reinforcement Learning Environment Library.

\bibitem[{Liu et~al.(2020)Liu, Yu, Feng, and Gao}]{IoTNetworksWithEdgeComputing}
Liu, X.; Yu, J.; Feng, Z.; and Gao, Y. 2020.
\newblock Multi-agent reinforcement learning for resource allocation in IoT networks with edge computing.
\newblock \emph{China Communications}, 17(9): 220--236.

\bibitem[{Lowe et~al.(2017)Lowe, Wu, Tamar, Harb, Abbeel, and Mordatch}]{lowe2017multi}
Lowe, R.; Wu, Y.; Tamar, A.; Harb, J.; Abbeel, P.; and Mordatch, I. 2017.
\newblock Multi-Agent Actor-Critic for Mixed Cooperative-Competitive Environments.
\newblock \emph{Neural Information Processing Systems (NIPS)}.

\bibitem[{Madumal et~al.(2020)Madumal, Miller, Sonenberg, and Vetere}]{madumal2020explainable}
Madumal, P.; Miller, T.; Sonenberg, L.; and Vetere, F. 2020.
\newblock Explainable reinforcement learning through a causal lens.
\newblock In \emph{Proceedings of the AAAI conference on artificial intelligence}, volume~34, 2493--2500.

\bibitem[{Mahajan et~al.(2020)Mahajan, Rashid, Samvelyan, and Whiteson}]{mahajan2020maven}
Mahajan, A.; Rashid, T.; Samvelyan, M.; and Whiteson, S. 2020.
\newblock MAVEN: Multi-Agent Variational Exploration.
\newblock arXiv:1910.07483.

\bibitem[{Mnih et~al.(2016)Mnih, Badia, Mirza, Graves, Lillicrap, Harley, Silver, and Kavukcuoglu}]{mnih2016asynchronous}
Mnih, V.; Badia, A.~P.; Mirza, M.; Graves, A.; Lillicrap, T.~P.; Harley, T.; Silver, D.; and Kavukcuoglu, K. 2016.
\newblock Asynchronous Methods for Deep Reinforcement Learning.
\newblock arXiv:1602.01783.

\bibitem[{Nasir and Guo(2019)}]{DynamicPowerAllocation}
Nasir, Y.~S.; and Guo, D. 2019.
\newblock Multi-Agent Deep Reinforcement Learning for Dynamic Power Allocation in Wireless Networks.
\newblock \emph{IEEE Journal on Selected Areas in Communications}, 37(10): 2239--2250.

\bibitem[{Papoudakis et~al.(2021)Papoudakis, Christianos, Schäfer, and Albrecht}]{papoudakis2021benchmarking}
Papoudakis, G.; Christianos, F.; Schäfer, L.; and Albrecht, S.~V. 2021.
\newblock Benchmarking Multi-Agent Deep Reinforcement Learning Algorithms in Cooperative Tasks.
\newblock arXiv:2006.07869.

\bibitem[{Pretorius et~al.(2020)Pretorius, Cameron, Van~Biljon, Makkink, Mawjee, du~Plessis, Shock, Laterre, and Beguir}]{pretorius2020game}
Pretorius, A.; Cameron, S.; Van~Biljon, E.; Makkink, T.; Mawjee, S.; du~Plessis, J.; Shock, J.; Laterre, A.; and Beguir, K. 2020.
\newblock A game-theoretic analysis of networked system control for common-pool resource management using multi-agent reinforcement learning.
\newblock \emph{Advances in Neural Information Processing Systems}, 33: 9983--9994.

\bibitem[{Puiutta and Veith(2020)}]{puiutta2020explainable}
Puiutta, E.; and Veith, E.~M. 2020.
\newblock Explainable reinforcement learning: A survey.
\newblock In \emph{Machine Learning and Knowledge Extraction: 4th IFIP TC 5, TC 12, WG 8.4, WG 8.9, WG 12.9 International Cross-Domain Conference, CD-MAKE 2020, Dublin, Ireland, August 25--28, 2020, Proceedings 4}, 77--95. Springer.

\bibitem[{Rashid et~al.(2020)Rashid, Farquhar, Peng, and Whiteson}]{rashid2020weighted}
Rashid, T.; Farquhar, G.; Peng, B.; and Whiteson, S. 2020.
\newblock Weighted QMIX: Expanding Monotonic Value Function Factorisation for Deep Multi-Agent Reinforcement Learning.
\newblock arXiv:2006.10800.

\bibitem[{Rashid et~al.(2018)Rashid, Samvelyan, de~Witt, Farquhar, Foerster, and Whiteson}]{rashid2018qmix}
Rashid, T.; Samvelyan, M.; de~Witt, C.~S.; Farquhar, G.; Foerster, J.; and Whiteson, S. 2018.
\newblock QMIX: Monotonic Value Function Factorisation for Deep Multi-Agent Reinforcement Learning.
\newblock arXiv:1803.11485.

\bibitem[{Samvelyan et~al.(2021)Samvelyan, Kirk, Kurin, Parker-Holder, Jiang, Hambro, Petroni, K{\"u}ttler, Grefenstette, and Rockt{\"a}schel}]{Samvelyan2021MiniHackTP}
Samvelyan, M.; Kirk, R.; Kurin, V.; Parker-Holder, J.; Jiang, M.; Hambro, E.; Petroni, F.; K{\"u}ttler, H.; Grefenstette, E.; and Rockt{\"a}schel, T. 2021.
\newblock MiniHack the Planet: A Sandbox for Open-Ended Reinforcement Learning Research.
\newblock \emph{ArXiv}, abs/2109.13202.

\bibitem[{Samvelyan et~al.(2019)Samvelyan, Rashid, de~Witt, Farquhar, Nardelli, Rudner, Hung, Torr, Foerster, and Whiteson}]{SMAC}
Samvelyan, M.; Rashid, T.; de~Witt, C.~S.; Farquhar, G.; Nardelli, N.; Rudner, T. G.~J.; Hung, C.-M.; Torr, P. H.~S.; Foerster, J.; and Whiteson, S. 2019.
\newblock The StarCraft Multi-Agent Challenge.

\bibitem[{Schulman et~al.(2017)Schulman, Wolski, Dhariwal, Radford, and Klimov}]{schulman2017proximal}
Schulman, J.; Wolski, F.; Dhariwal, P.; Radford, A.; and Klimov, O. 2017.
\newblock Proximal policy optimization algorithms.
\newblock \emph{arXiv preprint arXiv:1707.06347}.

\bibitem[{Spatharis et~al.(2019)Spatharis, Blekas, Bastas, Kravaris, and Vouros}]{AirTrafficManagement}
Spatharis, C.; Blekas, K.; Bastas, A.; Kravaris, T.; and Vouros, G.~A. 2019.
\newblock Collaborative multiagent reinforcement learning schemes for air traffic management.
\newblock In \emph{2019 10th International Conference on Information, Intelligence, Systems and Applications (IISA)}, 1--8.

\bibitem[{Sunehag et~al.(2017)Sunehag, Lever, Gruslys, Czarnecki, Zambaldi, Jaderberg, Lanctot, Sonnerat, Leibo, Tuyls, and Graepel}]{VDN}
Sunehag, P.; Lever, G.; Gruslys, A.; Czarnecki, W.~M.; Zambaldi, V.; Jaderberg, M.; Lanctot, M.; Sonnerat, N.; Leibo, J.~Z.; Tuyls, K.; and Graepel, T. 2017.
\newblock Value-Decomposition Networks For Cooperative Multi-Agent Learning.
\newblock arXiv:1706.05296.

\bibitem[{Tan(1993)}]{tan1993multi}
Tan, M. 1993.
\newblock Multi-agent reinforcement learning: Independent vs. cooperative agents.
\newblock In \emph{Proceedings of the tenth international conference on machine learning}, 330--337.

\bibitem[{Tan(1997)}]{IQL}
Tan, M. 1997.
\newblock Multi-Agent Reinforcement Learning: Independent versus Cooperative Agents.
\newblock In \emph{International Conference on Machine Learning}.

\bibitem[{Vidhate and Kulkarni(2017)}]{IntelligentTrafficControl}
Vidhate, D.~A.; and Kulkarni, P. 2017.
\newblock Cooperative multi-agent reinforcement learning models (CMRLM) for intelligent traffic control.
\newblock In \emph{2017 1st International Conference on Intelligent Systems and Information Management (ICISIM)}, 325--331.

\bibitem[{Vouros(2022)}]{vouros2022explainable}
Vouros, G.~A. 2022.
\newblock Explainable deep reinforcement learning: state of the art and challenges.
\newblock \emph{ACM Computing Surveys}, 55(5): 1--39.

\bibitem[{Wang et~al.(2021)Wang, Ren, Liu, Yu, and Zhang}]{wang2021qplex}
Wang, J.; Ren, Z.; Liu, T.; Yu, Y.; and Zhang, C. 2021.
\newblock QPLEX: Duplex Dueling Multi-Agent Q-Learning.
\newblock arXiv:2008.01062.

\bibitem[{Wang et~al.(2020)Wang, Gupta, Mahajan, Peng, Whiteson, and Zhang}]{wang2020rode}
Wang, T.; Gupta, T.; Mahajan, A.; Peng, B.; Whiteson, S.; and Zhang, C. 2020.
\newblock RODE: Learning Roles to Decompose Multi-Agent Tasks.
\newblock arXiv:2010.01523.

\bibitem[{Yu et~al.(2021)Yu, Velu, Vinitsky, Wang, Bayen, and Wu}]{yu2021}
Yu, C.; Velu, A.; Vinitsky, E.; Wang, Y.; Bayen, A.; and Wu, Y. 2021.
\newblock The Surprising Effectiveness of PPO in Cooperative, Multi-Agent Games.

\bibitem[{Zhao, Liu, and Cheng(2020)}]{TrajectoryDesignandPowerAllocation}
Zhao, N.; Liu, Z.; and Cheng, Y. 2020.
\newblock Multi-Agent Deep Reinforcement Learning for Trajectory Design and Power Allocation in Multi-UAV Networks.
\newblock \emph{IEEE Access}, 8: 139670--139679.

\end{thebibliography}
